\documentclass[11pt]{article}
\usepackage{acl2015}
\usepackage{times}
\usepackage{url}
\usepackage{latexsym}
\usepackage[dvipsnames]{xcolor}
\usepackage{graphicx}
\usepackage{bm}
\definecolor{paperBLUE}{RGB}{5, 6, 160}
\usepackage[colorlinks=true,linkcolor=paperBLUE,citecolor=paperBLUE, allcolors=paperBLUE, hypertexnames=true]{hyperref}%

\usepackage{booktabs} 
\usepackage{lipsum}
\usepackage[all]{hypcap}



\usepackage{natbib}
\setcitestyle{authoryear}
\setcitestyle{notesep={ -}}

\title{Long-length Legal Document Classification \vspace{0.5cm}}

\author{\textbf{Lulu Wan} \\
  Clifford Chance LLP\\
  {\tt {\fontsize{9.5}{12}\selectfont lulu.wan@cliffordchance.com}} \\ \\
  \textbf{Michael Seddon} \\
  Clifford Chance LLP \\ 
  {\tt {\fontsize{9.4}{12}\selectfont michael.seddon@cliffordchance.com}} \\ \hspace{5cm}\And
  \textbf{George Papageorgiou} \\
  Clifford Chance LLP\\
  {\tt {\fontsize{9.5}{12}\selectfont george.papageorgiou@cliffordchance.com}} \\ \\
  \textbf{Mirko Bernardoni} \\
  Clifford Chance LLP \\
  {\tt {\fontsize{9.4}{12}\selectfont mirko.bernardoni@cliffordchance.com}}  \\}

\vspace{2cm}

\date{}

\begin{document}

\maketitle

\begin{abstract}
 One of the principal tasks of machine learning with major applications is text classification. This paper focuses on the legal domain and, in particular, on the classification of lengthy legal documents. The main challenge that this study addresses is the limitation that current models impose on the length of the input text. In addition, the present paper shows that dividing the text into segments and later combining the resulting embeddings with a BiLSTM architecture to form a single document embedding can improve results. These advancements are achieved by utilising a simpler structure, rather than an increasingly complex one, which is often the case in NLP research. The dataset used in this paper is obtained from an online public database containing lengthy legal documents with highly domain-specific vocabulary and thus, the comparison of our results to the ones produced by models implemented on the commonly used datasets would be unjustified. This work provides the foundation for future work in document classification in the legal field.
\end{abstract}

\section{Introduction}

Text classification is a problem in library, information and computer science and one of the most classical and prominent tasks in Natural Language Processing (NLP). In particular, document classification is a procedure of assigning one or more labels to a document from a predetermined set of labels. Automatic document classification tasks can be divided into three categories: supervised, unsupervised and semi-supervised. This study focuses on supervised document classification.

Research so far has focused on short text \citep{yang2016hierarchical, category, zhang2019mitigating, docbert}, whereas the main objective of this paper is to address the classification of lengthy legal documents. In fact, pre-existing models could not be applied on our corpus, which consists of excessively lengthy legal documents. In the legal field, companies manage millions of documents per year, depending on the size of the company. Therefore, automatic categorisation of documents into different groups significantly enhances the efficiency of document management and decreases the time spent by legal experts analysing documents.

Recently, several quite sophisticated frameworks have been proposed to address the document classification task. However, as proven by \citet{docbert} regarding the document classification task, complex neural networks such as Bidirectional Encoder Representations from Transformers (BERT; \citealp{1810.bert}) can be distilled and yet achieve similar performance scores. In addition, \citet{rethinkingcomplexity} shows that complex architectures are more sensitive to hyperparameter fluctuations and are susceptible to domains that consist of data with dissimilar characteristics. In this study, rather than employing an overly complex neural architecture, we focus on a relatively simpler neural structure that, in short, creates text embeddings using Doc2Vec \citep{Doc2vec} and then passes them through a Bi-directional LSTM (BiLSTM) with attention before making the final prediction.    

Furthermore, an important contribution of this paper to automatic document classification is the concept of dividing documents into chunks before processing. It is demonstrated that the segmentation of lengthy documents into smaller chunks of text allows the context of each document to be encapsulated in an improved way, leading to enhanced results. The intuition behind this idea was formed by investigating automatic audio segmentation research. Audio segmentation (also known as audio classification) is an essential pre-processing step in audio analysis that separates different types of sound (e.g. speech, music, silence etc.) and splits audio signals into chunks in order to further improve the comprehension of these signals \citep{IntroToAudio}. Analogously, the present paper shows that splitting overly lengthy legal documents into smaller parts before processing them, boosts the final results.

\begin{table}[]
    \centering
    \begin{tabular}{lrrrr} \toprule
     \textbf{Dataset} &  \textbf{C} & \textbf{N} & \textbf{W} &\textbf{S}  \\\hline
     Reuters & 90 & 10,789 & 144.3 & 6.6  \\
     AAPD & 54 & 55,840 & 167.3 & 1.0 \\
     IMDB &  10 &135,669 & 393.8 & 14.4  \\
     Yelp 2014 & 5 & 1,125,386 & 148.8 & 9.1  \\ \bottomrule
    \end{tabular}
    \caption{Summary of the commonly used datasets for document classification. C denotes the number of classes in the dataset, N the number of samples and W and S the average number of words and sentences per document respectively.}
    \label{table_otherdataset}
\end{table}


\section{Related Work}
In several industries that produce or handle colossal amounts of text data such as the legal industry, document categorisation is still often performed manually by human experts. Automatic categorisation of documents is highly beneficial for reducing the human effort spent on time-consuming operations. In particular, deep neural networks have achieved state-of-the-art results in document classification over the last few years, outperforming the human classifiers in numerous cases.

\subsection{Document Classification Datasets}
The majority of researchers evaluate their document classifying models on the following four datasets: Reuters-21578 \citep{Reuters21578}, ArXiv Academic Paper Dataset - AAPD \citep{aapd}, IMDB reviews \citep{imdb}, and Yelp 2014 reviews \citep{yelp}. However, these commonly used datasets do not contain large documents, which conflicts with one of the main objectives of this study. Note that our definition of `document' in this specific context is a document that has at least 5000 words.

For that purpose, we use a  dataset  provided  by  the  U.S  Securities  and  Exchange  Commission  (SEC), namely EDGAR \citep[- see Section~\ref{Dataset}]{edgar}. As anticipated, most models that have achieved inspiring results have very poor performance or even fail when they are tested on large documents from the EDGAR corpus. 
As shown in Table~\ref{table_otherdataset} and Table~\ref{table_dataset}, the differences between the commonly used datasets and the EDGAR dataset are evident.

\begin{table}[] 
    \centering
    \begin{tabular}{lrrr} \toprule
     \textbf{Type} &  \textbf{N} & \textbf{W} & \textbf{S}  \\\hline
     EX-101.INS & 5,689 & 11,730 & 344  \\
     EX-10.1 &  5,689 & 28,515 & 432 \\
     10-Q &  5,689 & 20,178 & 552  \\
     EX-99.1 & 5,689 & 12,224 & 276  \\
     10-K & 5,689 &  51,071 & 1,476 \\ \bottomrule
    \end{tabular}
    \caption{Summary of EDGAR dataset. N denotes the number of samples and W and S the average number of words and sentences per document respectively.}
    \label{table_dataset}
\end{table}

\subsection{Document Classification Approaches}

The application of deep neural networks in the field of computer vision has achieved great success. Following this success, several well-known DNN models attained remarkable results when applied on the document classification task. One of the most popular models is the Hierarchical Attention Network (HAN) proposed by \citet{yang2016hierarchical}. HAN used word and sentence-level attention in order to extract meaningful features of the documents and ultimately classify them. However, the fact that this architecture is based on a Gated Recurrent Unit (GRU) framework combined with the excessive size of the documents in our corpus would severely affect the results. Concretely, using overly large documents would result in a vast number of time steps and the vanishing gradient problem would be detrimental to performance.

A different yet powerful framework, namely BERT \citep{1810.bert}, has achieved state-of-the art results on a large amount of NLP tasks. BERT architecture employs self-attention instead of general attention, thus making the neural network even more complex. Nevertheless, \citet{docbert} have established groundbreaking results and demonstrated that sophisticated architectures such as BERT are not necessary to succeed in the document classification task.  
Furthermore, it is worth mentioning that both the aforementioned models were trained on a rather different corpora. The main difference between the datasets used by those researchers and the EDGAR dataset is the size of the documents, which explains why these models could not be utilised in the present study. In particular, BERT was incompatible with our dataset due to the maximum input sequence length that imposes, namely the 512 terms threshold. 


\section{Methods}

The novelty of this work is the application of audio segmentation used for speech recognition \citep{audio_segmentation} in document classification. The ultimate purpose of audio segmentation is to divide the signal into segments, each of which contains distinct audio information. In our case, the same occurs during the document segmentation, where the split chunks become the inputs of our neural network.

\begin{figure}[th]
\centering
\includegraphics[width=5.6cm, height=9.5cm]{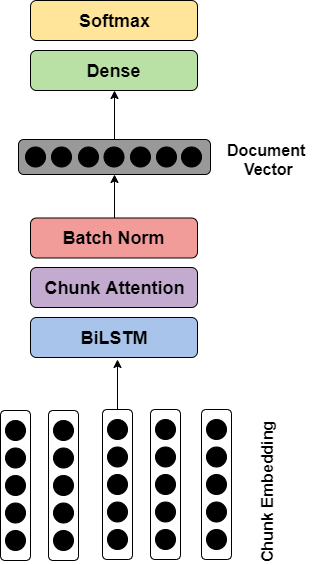}
\caption{Overall architecture of proposed BiLSTM model.}
\label{fig:model}
\end{figure}

From a human perspective, when reading a rather long document or book, we are constantly storing and updating our memory with the essential parts or information of that record. Once enough information is stored in our memory we can form connections so as to gain a deeper understanding of the context and potentially extract valuable insight. In the same way, instead of passing the whole document to Doc2Vec, we split the document into multiple chunks (Figure~\ref{fig:model}). Hence, the machine can imitate human behaviour by identifying and determining the relevance of each chunk.

We create different models with respect to the number of chunks that we divide the initial text into, in order to observe how the different number of chunks affect the efficiency of the final model. These chunks are then used to train Doc2Vec.
In short, the intuition behind Doc2Vec is analogous to the intuition behind Word2Vec, where the 

words are used to make predictions about the target word (central word). The additional part of Doc2Vec is that it also considers the document ID when predicting a word. Ultimately, after the training each chunk has the form of an embedding.    

\begin{figure}[t]
\includegraphics[width=7.9cm, height=6cm]{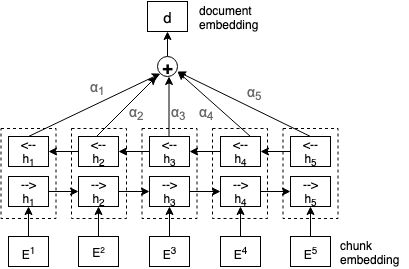}
\caption{Document embedding process through BiLSTM framework.}
\label{fig:BiLSTM}
\end{figure}

In the next phase, we aggregate the different chunk embeddings of a document into one vector through the use of a BiLSTM (see Figure~\ref{fig:BiLSTM}). First, the different chunk embeddings $E_{i}^1, E_{i}^2,..., E_{i}^n$ of a document are sequentially fed to the BiLSTM model. Then, the outputs of the forward and the backward layer are concatenated; $h_{it}=[\overrightarrow{h_{it}}\overleftarrow{h_{it}}]$. $h_{it}$ denotes the resulting vectors. 

The final classification is subjected to the various features that each chunk contains. Thus, the attention mechanisms are introduced so as to enable the assignment of different weights to each chunk, depending on how strong of a class indicator this chunk is. In particular, the attention scores are assigned to the corresponding hidden state outputs as follows:

\begin{equation}
    u_{it}=tanh(W^{a}h_{it}+b^a)
\end{equation}    
\begin{equation}
    \alpha_{it}=\frac{exp(u_{it}^{T}u_{w})}{\sum_{t}exp(u_{it}^{T}u_{w})}
\end{equation}    
\begin{equation}
    d_{i}=\sum \alpha_{it}h_{it}
\end{equation}

Here $\alpha_{it}$ is the attention score assigned to hidden state $h_{it}$ of document $i$ at time step $t$. This score is determined by the similarity between $u_{it}$ and $u_{w}$, where $u_{it}$ is a mere non-linear transformation of $h_{it}$ and $u_{w}$ is the context (category) vector \citep{category}. During the following steps, the products of the hidden states and their corresponding attention scores are calculated and the document vector $d_{i}$ is formed from the summation of those products. Note that $u_{w}$ is randomly initialised and then constantly updated during the training process.

Ultimately, we try different classifiers in order to assess the impact of the segmentation method. As part of the models of the first type, the resulting document vector is output from a batch normalisation layer. A linear transformation is then applied to that and this output is passed through a softmax classifier in order to acquire the multi-class probabilities. This final process is summarised in the following formula:  
\begin{equation}
    s_{i}=softmax(BN(Wd_{i}+b))
\end{equation}
where $W\in R^{c\times d}$ is the weight matrix, $c$ and $d$ are the number of classes and the number of dimensions of the hidden states respectively and $b\in R^d$ is the bias term. Hence, the final vector $s_{i}$ is a c-dimension vector comprising the probability of that document belonging to each class. 

The models of the second type are based on a strong machine learning classifier, namely Support Vector Machine (SVM). SVM also performs document classification by utilising the resulting document embeddings. The main parameters used to train SVM were obtained by optimising each model separately (see Section~\ref{model_conf}).

\section{Experimental Setup}

We evaluate the proposed model on a document classification dataset; 70\% of the data is used for the training and the remaining 30\% is equally divided and used for tuning and testing our model.

During the pre-processing stage where the documents are split into chunks, we utilise a cluster of Azure Virtual Machines with 32 GB RAM and 16 cores, which are optimised for CPU usage. A similar cluster is used during the hyperparameter optimisation, however, with 112 GB RAM. Reading from the remote disk (Azure Blob Storage) is rather time-consuming, since the corpus comprises lengthy documents. Thus, to accelerate the training, we chose nodes with abundant memory in order to load everything in memory just once (required roughly one hour for that process).  

We use Pytorch 1.2.0 as the backend framework, Scikit-learn 0.20.3 for SVM and dataset splits, and gensim 3.8.1 for Doc2Vec model.

\begin{table}[t] 
    \centering
    \begin{tabular}{lp{4.5cm}}\toprule
    \textbf{Filing Type} & \textbf{Description under Regulation S-K \footnotemark} \\ \hline
      10-Q           &  Quarterly reports \\
      10-K           &  Annual reports \\
      EX-10.1        &  Material contracts  \\ 
      EX-99.1        &  Additional exhibits for investment companies \\ 
      EX-101.INS     &  XBRL-related documents \\ \bottomrule
    \end{tabular}
    \caption{Description of different filing type contents.}
    \label{tab:selected_types}
\end{table}

\footnotetext{Regulation S-K is an official regulation under the US Securities Act of 1933 that establishes reporting regulations for a variety SEC filings used by public companies.}

\subsection{Dataset} \label{Dataset}
The data we use to evaluate our model is a set of documents downloaded from EDGAR, an online public database from the U.S. Securities and Exchange Commission (SEC). EDGAR is the primary system for submissions by companies and others who are required by law to file information with the SEC. 
These documents can be grouped according to filing types, which determines the substantial content to fulfill their filing obligation. 
To work on as many documents as possible, we choose the following types: ``10-Q'', ``10-K'', ``EX-99.1'', ``EX-10.1'' and ``EX-101.INS''.
The total number of documents is 28,445 and there are 5,689 documents for each filing type.
We summarise the statistics of this dataset in Table~\ref{tab:selected_types}. 

Almost all documents of type ``10-K'' begin with lines that contain identical headings. In order to enable the machine to truly comprehend why a document of type ``10-K'' should be categorised to that filing type, we remove the first six lines where the identical text is located. The model is then able to focus on finding common features that exist in documents of the same filing type, rather than focusing on just capturing the few sentences that are the same in almost all of the documents of type ``10-K''. A similar procedure is followed with the documents of type ``10-Q''.

\begin{table}[t]
    \centering
    \begin{tabular}{lrrr} \toprule
     \textbf{Model} & \bm{$W_{c}$\ \ } & \textbf{Test.F1} & \textbf{Val.F1}  \\\hline
     1-chunk &  24,744  & 96.96 & 97.50 \\
     3-chunk &  8,248    & 97.85 & \textbf{98.11} \\
     5-chunk &  4,949    & \textbf{97.97} & 97.85 \\
     7-chunk &  3,535    & 97.45 & 97.55 \\
     10-chunk & 2,474  & 97.87 & 97.87 \\
     25-chunk & 990   & 97.41 & 97.64 \\
     50-chunk & 495   & 97.34 & 97.22 \\
    \bottomrule
    \end{tabular}
    \caption{Performance of models of the first type (simple linear classifier) reported on validation and test set. $W_{c}$ denotes the average words per chunk and best scores are shown in bold.}
    \label{table_results_BiLSTM}
\end{table}

\subsection{Model Configuration} \label{model_conf}

As Table~\ref{table_results_BiLSTM} shows, we create seven different models that correspond to the number of chunks that the text is divided into before passing through Doc2Vec. Each model is optimised separately to ensure fair comparison. 
 
For the optimisation of the BiLSTM with attention model, we use Adam optimiser with a learning rate of 0.001, batch size of 1,000 and distinct values for each one of the other hyperparameters. Analogously, the SVM classifier consists of the Radial Basis Function (RBF) as the kernel function and a different value of gamma and the penalty parameter for each different model. The intention of the distinct values used for each model is to optimise each model separately so as to enable them to reach their best performance.  

Furthermore, we observe that Doc2Vec requires only a small portion of the corpus to train accurately. Indeed, when training Doc2Vec on more documents we observe a substantial decrease in accuracy. It is well-known that legal documents contain several domain-specific words that are often repeated not only among different documents, but also within the same document. Training Doc2Vec on more documents introduced undesirable noise that results from company names, numbers such as transaction amounts and dates, job titles and addresses. Consequently, Doc2Vec is proven to generate more accurate document embeddings when trained on just 150 randomly chosen documents (30 for each filing type). 
 
\begin{table}[t]
    \centering
    \begin{tabular}{lrrr} \toprule
     \textbf{Model} & \bm{$W_{c}$\ \ } & \textbf{Test.F1} & \textbf{Val.F1}  \\\hline
     1-chunk &  24,744 & 97.71 & 97.50 \\
     3-chunk &  8,248  & 97.97 & \textbf{98.20} \\
     5-chunk &  4,949  & 98.04 & 98.04 \\
     7-chunk &  3,535  & \textbf{98.11} & 97.83 \\
     10-chunk & 2,474  & 97.92 & 97.83 \\
     25-chunk & 990    & 97.64 & 97.92 \\
     50-chunk & 495    & 97.24 & 97.38 \\
    \bottomrule
    \end{tabular}
    \caption{Performance of models of the second type (SVM classifier) reported on validation and test set. $W_{c}$ denotes the average words per chunk and best scores are shown in bold.}
    \label{table_results_+SVM}
\end{table}

\begin{figure*}[h]
\includegraphics[width=\textwidth]{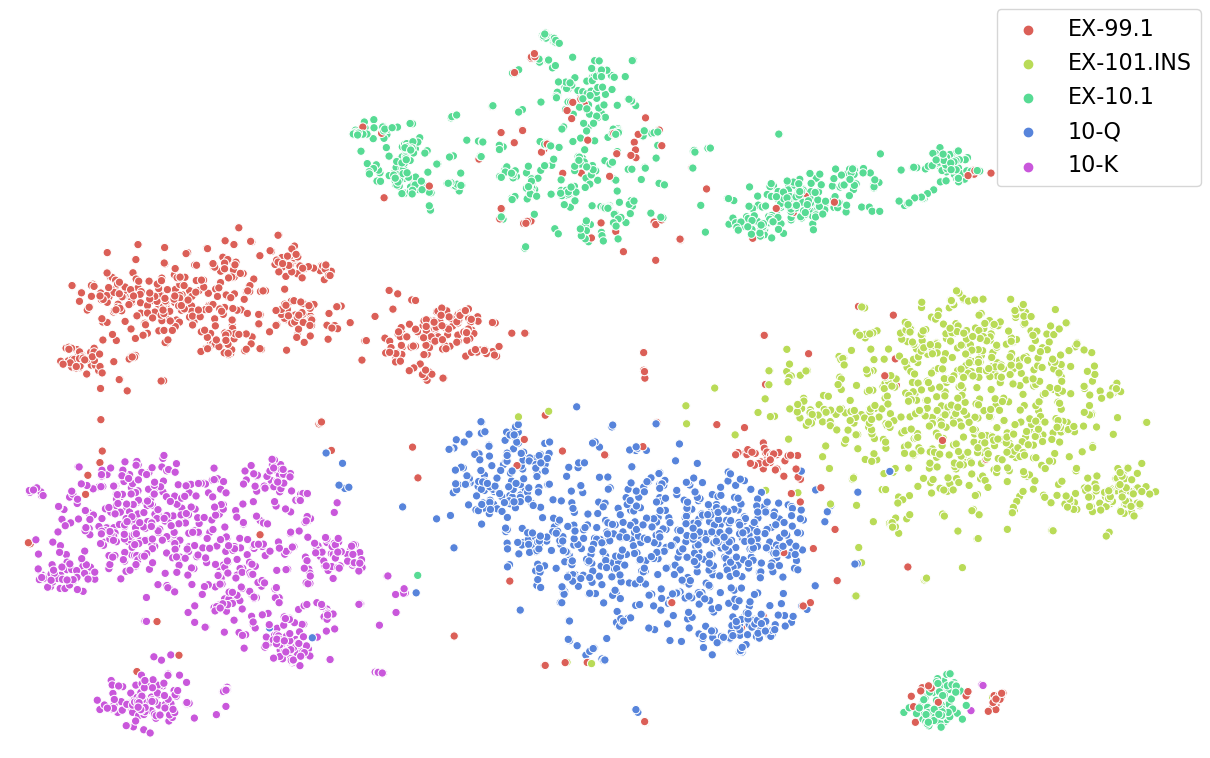}
\caption{t-SNE plot of projections of document embeddings, using vanilla Doc2Vec.}
\label{fig:TSNE_3}
\end{figure*}

\begin{figure*}[!h]
\includegraphics[width=\textwidth]{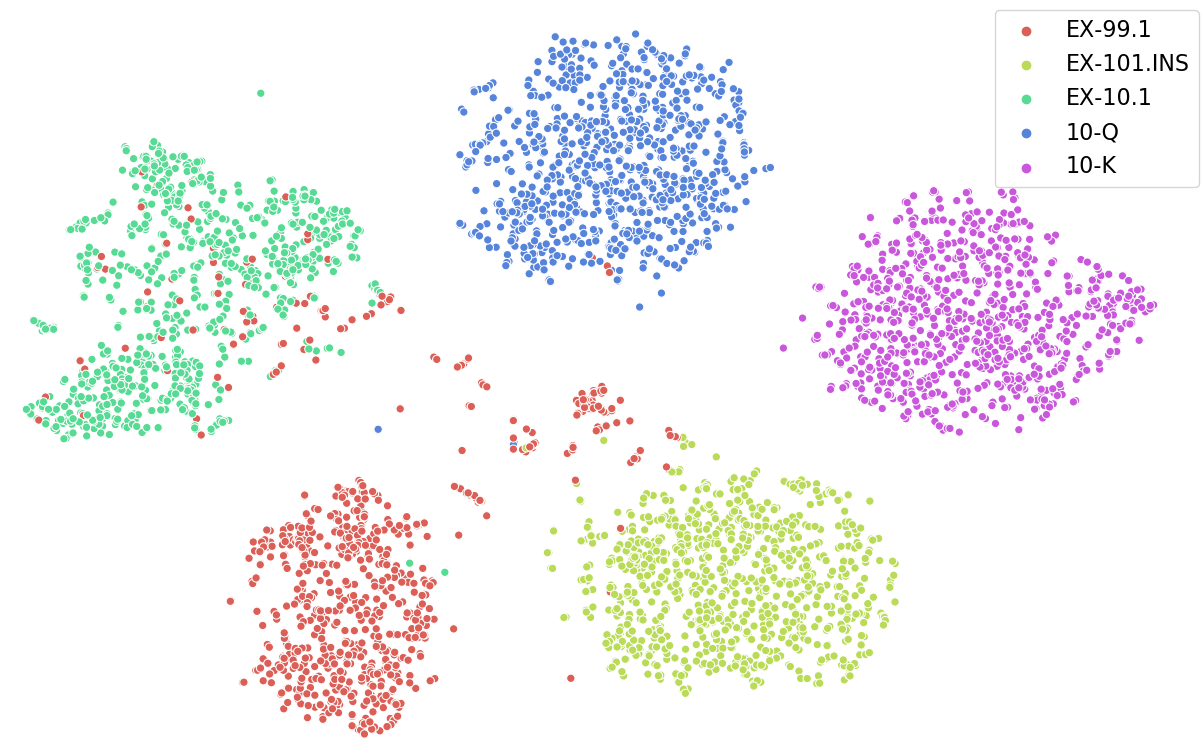}
\caption{t-SNE plot of projections of document embeddings, using Doc2Vec + BiLSTM.}
\label{fig:TSNE_test}
\end{figure*}


\section{Results and Discussion}

\begin{figure}[th]
\includegraphics[width=8.6cm, height=8.5cm]{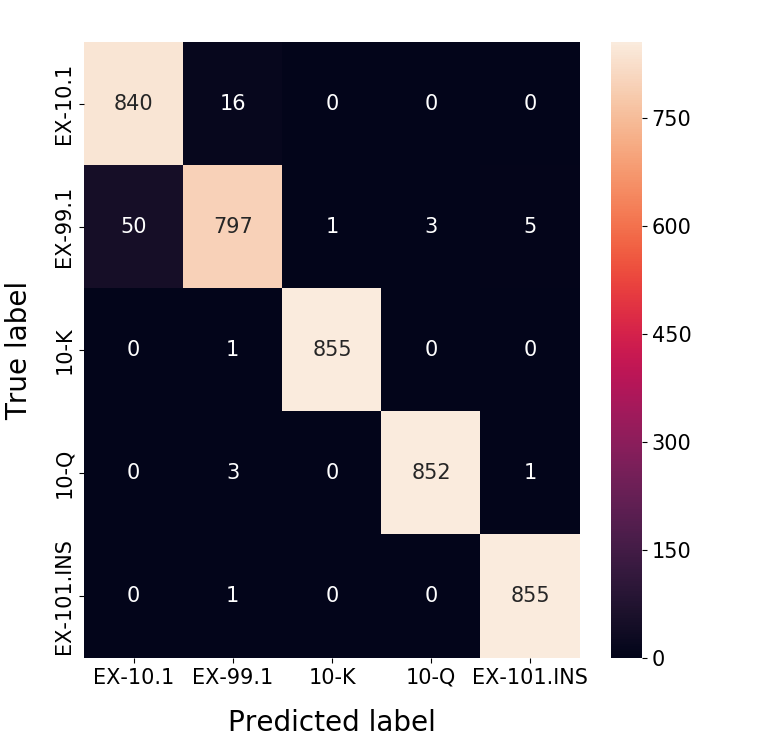}
\caption{Confusion matrix plot of classification results for 7-chunk model on test set.}
\label{fig:conf_matrix}
\end{figure}

Recently, reproducibility is becoming a growing concern for the NLP community \citep{crane2018questionable}. In fact, the majority of the papers we consider in this study fail to report the validation set results. To address these issues, apart from the F1 scores on the test sets we also report the F1 scores for the validation sets.

Legal documents contain domain-specific vocabulary and each type of document is normally defined in a very unambiguous way. Hence, even simple classifiers can achieve relatively high accuracy when classifying different documents. Nevertheless, even the slightest improvement of 1\% or less will result in the correct classification of thousands of additional documents, which is crucial in the legal industry when handling large numbers of documents. This research allows these simple classifiers to achieve even greater results, by combining them with different architectures. 

As Table~\ref{table_results_BiLSTM} and Table~\ref{table_results_+SVM} indicate, dividing the document in chunks - up to certain thresholds - results in improved models compared to those where the whole document is input into the classifier. Note that the model with one chunk denotes the model which takes as input the whole document to produce the document embedding and thereby is used as a benchmark in order to be able to identify the effectiveness of the document segmentation method. 

More specifically, splitting the document into chunks yields higher test accuracy than having the whole document as input. Our first model with the BiLSTM based framework and the linear classifier reaches a 97.97\% accuracy with a 1.1\% improvement upon the benchmark model. Similarly, the second model with the SVM classifier reaches a remarkable 98.11\% accuracy with a 0.4\% improvement upon the benchmark model.

A more thorough investigation of the test accuracy scores indicate that documents of type ``EX-99.1" are the ones that get misclassified the most, whereas the remaining four types of documents are in general classified correctly at a considerably higher rate. As confusion matrix plot in Figure~\ref{fig:conf_matrix} highlights, there are cases that documents of type ``EX-10.1" are misclassified as ``EX-99.1", however, the reverse occurs more frequently. Further exploration of documents of type ``EX-99.1" reveals that these documents often contain homogeneous agreements or clauses with the ones embodied in documents of type ``EX-10.1". 

Ultimately, Figure~\ref{fig:TSNE_3} and Figure~\ref{fig:TSNE_test} demonstrate the increase of the efficiency of the document embeddings after the use of BiLSTM. These vector representations of each cluster have noticeably more robustly defined boundaries after they are passed through the BiLSTM network compared to the ones that are only passed through the mere Doc2Vec. 


\section{Conclusion}

The main contribution of this paper is to overcome the document length limitations that are imposed by most modern architectures. It also shows that dividing documents into chunks before inputting them into Doc2Vec can result in enhanced models. Nonetheless, these advancements are accomplished with a relatively simplified structure, rather than a significantly more sophisticated architecture than its predecessors, which is often the case in NLP research.

One potential extension of this work would be to apply powerful yet computationally expensive pre-processing techniques to the various documents. Techniques such as Named Entity Recognition (NER) could enable the training of the whole corpus in Doc2Vec by removing the undesired noise. Furthermore, the projections of the document embeddings at the end of our pipeline are shown to have clearly defined boundaries and thus they can be valuable for different NLP tasks, such as estimating document similarities. In the legal industry, this can contribute to identifying usages of legal templates and clauses.


\bibliography{ref}

\end{document}